\documentclass[a4paper,twoside]{article}

\usepackage{epsfig}
\usepackage{subcaption}
\usepackage{calc}
\usepackage{amssymb}
\usepackage{amstext}
\usepackage{amsmath}
\usepackage{amsthm}
\usepackage{multicol}
\usepackage{pslatex}
\usepackage{apalike}
\usepackage{algorithm2e}
\usepackage[bottom]{footmisc}
\usepackage{SCITEPRESS}     

\usepackage{multirow}
\usepackage{booktabs} 
\usepackage{graphicx} 
\usepackage{stfloats}
\usepackage{float}
\usepackage{cite}
\usepackage{amsfonts}
\usepackage{algorithmic}
\usepackage{textcomp}
\usepackage{xcolor}
\usepackage{url}

\begin{document}

\title{FlowAct: A Proactive Multimodal Human-robot Interaction System with Continuous Flow of Perception and Modular Action Sub-systems}

 \author{\authorname{Timothée Dhaussy\sup{1}, Bassam Jabaian\sup{1} and Fabrice Lefèvre\sup{1}}
 \affiliation{\sup{1}Laboratoire Informatique d'Avignon, Avignon University, France}
 \email{\{timothee.dhaussy, bassam.jabaian, fabrice.lefevre\}@univ-avignon.fr}
 }


\abstract{The evolution of autonomous systems in the context of human-robot interaction systems requires a synergy between the continuous perception of the environment and the potential actions to navigate or interact with it. In this paper we present FlowAct, a proactive multimodal human-robot interaction architecture, working as an asynchronous endless loop of robot sensors into actuators, and organized by two controllers, the Environment State Tracking (EST) and the Action Planner. 
Through a series of real-world experiments, we exhibit the efficacy of the system in maintaining a continuous perception-action loop, substantially enhancing the responsiveness and adaptability of autonomous pro-active agents. The modular architecture of the action subsystems facilitates easy extensibility and adaptability to a broad spectrum of tasks and scenarios. The experiments demonstrate the ability of a Pepper robot governed by FlowAct to intervene proactively in laboratory tests and in the field in a hospital waiting room to offer participants various services (appointment management, information, entertainment, etc.).}

\onecolumn \maketitle \normalsize \setcounter{footnote}{0} \vfill

\section{\uppercase{Introduction}}
\label{sec:introduction}
Human-robot Interaction (HRI) has undergone a transformative journey, evolving from basic task-oriented engagements to sophisticated context-based interactions that mirror human-like dynamism~\cite{industryrob_grau_2021}. As robots become an integral part of our daily environments, there is a growing demand for systems that can continuously perceive, comprehend, and act within their surroundings in a way that is both intuitive and adaptive~\cite{cognitivecomputing_chen_2018}. The concept of continuous perception, where an autonomous entity perpetually senses and interprets its environment, has become a fundamental pillar for modern HRI systems~\cite{neuralcontrol_salomon_1997}. This is a departure from traditional systems that operated predominantly in a reactive mode, responding to stimuli based on preset algorithms or rules. 

The compartmentalization of specific functionalities into distinct modules, such as those for movement or speech, has been recognized as a crucial advance in the field~\cite{autonomous_seq_tekulve_2019}. This modular approach not only ensures that the system remains relevant in various scenarios, but also facilitates the seamless integration of new functionalities and the optimization of existing ones~\cite{teachproactive_garrell_2017}.

In this paper, we present FlowAct, a proactive multimodal system that exemplifies the fusion of continuous perception with action planning and monitoring. Anchored in the Environment State Tracker, FlowAct offers a representation of its surroundings, setting the stage for more informed and dynamic interactions, which can be proactively triggered.

\section{\uppercase{Related work}}
%
In recent decades, interactive robots designed to interact with humans have found widespread applications in diverse sectors. They are increasingly being used in service-orientated roles, such as serving as waiters in restaurants~\cite{hri_restaurant_2023}, working as customer guides in shopping malls~\cite{hri_mall_kanda_2009}, or assisting passengers in train stations~\cite{hri_train_shiomi_2011} to name a few. Moreover, these systems have made significant inroads in the healthcare sector~\cite{hri_hospital_2011, hri_hospital_2014, hri_hospital_2018}.

To build an HRI system, the architecture should integrate various software components to facilitate efficient and concurrent execution of multiple tasks. These systems must possess key capabilities, including recording historical events~\cite{hri_memory_2019}, and constructing representations of others' actions, beliefs, desires, and intentions~\cite{bdi_mccann_1991}. In their paper, Moulin-Firer \textit{et al.}~\cite{hri_dac3_2018} propose a cognitive architecture organization based on Distributed Adaptive Control (DAC)~\cite{dac_2003,dac_2014} that deals with the processing of states of the world, or exteroception, the self, or interception, and action. Furthermore, the RoboCog model of the ADAPTA project~\cite{hri_cortex_2015} enabled a sales robot to persuade potential customers to approach a sales booth. This robot was able to identify customers, gauge their willingness to follow, and respond to specific queries. 

The BRILLO~\cite{hri_brillo_2022} architecture for a bartender social robot follows a three-layer organization for its architecture: the execution layer, context awareness and decision-making layer, and percepts layer, all implemented within the Robot Operating System (ROS), an efficient software with libraries and tools to build robot applications. Although the architectures described above are specifically designed and adapted to a particular application task, they share a common processing structure with three layers: the perception layer, the representation layer, and the action layer. Our model was developed with this foundational concept in mind. Nevertheless, it was designed to be entirely task-independent, modular, and orientated towards process flow. Consequently, the stream of multimodal perception-actions is continuously processed, facilitating proactive action decisions through persistent monitoring of the perceived environment.

\section{\uppercase{FlowAct model}}
%
\subsection{Overview}
%
In this section, the specific features of FlowAct, a system designed to serve as a continuous conduit for sensory inputs and their resulting actions, are introduced. The next four sections provide a more precise insight into the implementation of the FlowAct layers.

\begin{figure}[t!]
  \hspace{-0.5cm}
  \centering
  \includegraphics[scale=0.17]{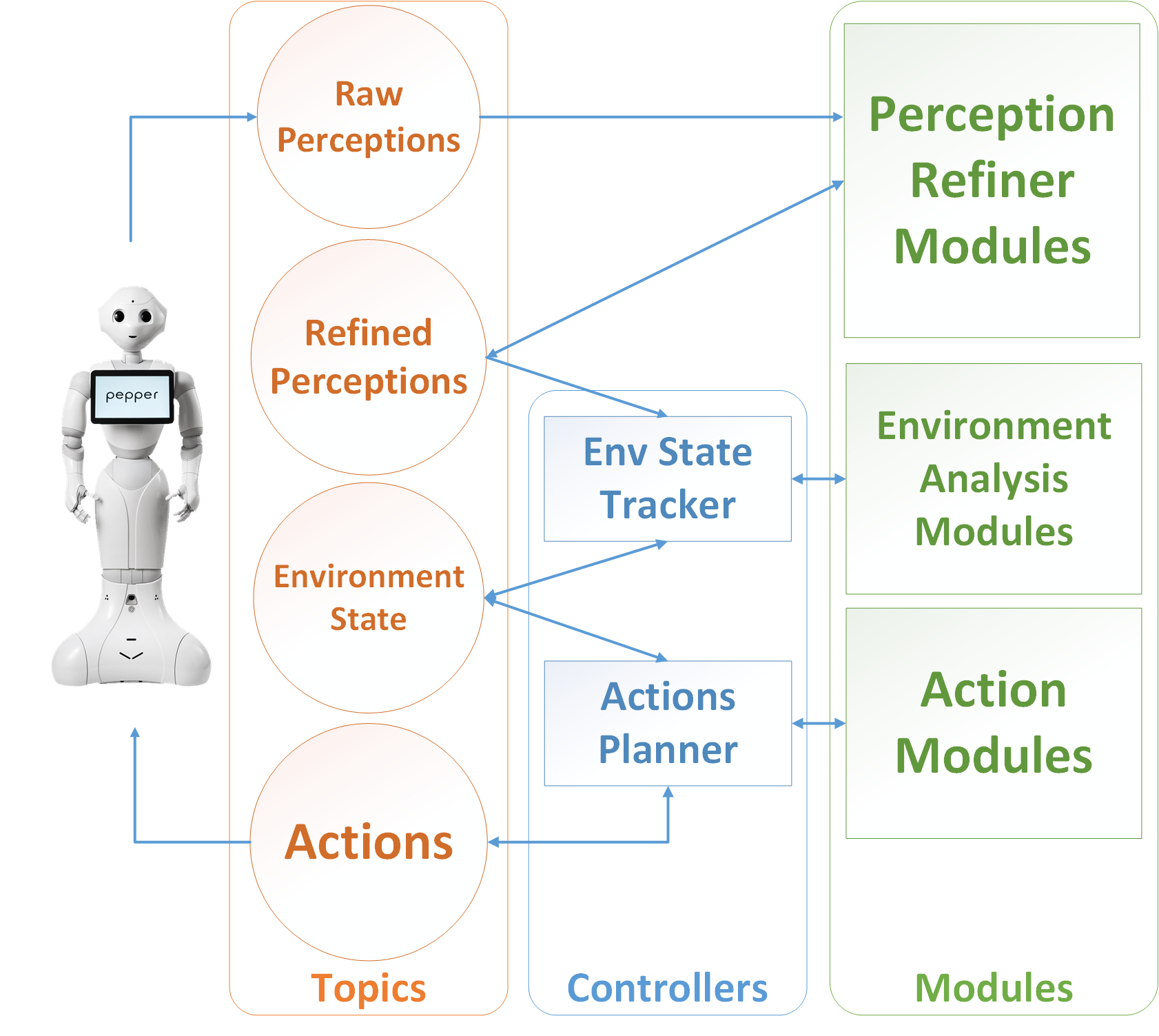}
  \caption{Overview of the FlowAct architecture, a continuous perception/action loop.}
  \label{figure:flowact}
\end{figure}

In FlowAct, the agent interacts with the world by performing an infinite loop, with sensors providing inputs and actuators delivering outputs. The cognitive architecture comprises three stages: perceiving the world through visual and audio sensors, representing the internal scene for the agent, and making decisions of actions to act on the world (move, speak, touch, etc.), as illustrated in Figure~\ref{figure:flowact}. As such, this overall structure follows the standard loop of cognitive architectures (perception, representation, and action)~\cite{hri_dac3_2018, hri_brillo_2022}. Moreover, a modular approach is implemented in which distinct modules are connected to controllers that perform essential functions within the architecture. This configuration facilitates the adaptation of modules to meet any particular cognitive requirements. The information requisite and disseminated at each stage is transmitted via dedicated memory zones or blackboards (referred to as "topics"), rendering it accessible to all controllers and permitting stringent regulation of production and consumption conditions, such as creation and modification timestamps, as well as access priority lists.

FlowAct separates the perception layer into two sub-levels, "raw perceptions" and "refined perceptions". The concept of raw perception refers to the agent's raw sensory data reflecting the quintet of human senses, especially auditory and visual, as well as perceptions directly derived from these senses, such as depth vision. These elementary perceptions are intercepted by the "perception refinement modules," a set in which raw data are distilled, producing an enhanced interpretative layer of the environment, called "refined perceptions."

The EST controller functions continuously, assimilating raw and refined perceptions to build a dynamic representation of the environment. It is also tasked with memory management and can interface with a database to store or retrieve specific knowledge, helping to interpret the current state of the world, such as the tracking of individuals or objects. This controller employs various environmental modules, including the re-identification of individuals and the allocation of perceptions, to update the scene analysis while taking into account its historical context.

The environment state is available to the Action Planner controller, the strategic core of FlowAct. This module is responsible for the analysis of environmental data and the planning of action strategies, informing the action modules asynchronously. It has the unique ability to reflect on future actions while executing current actions. The action modules, acting as the actuators of the system, transmit precise behavioral directives to the robotic agent, thus realizing the transition from the environmental state to action within this autonomous loop.

\subsection{Robot perceptions}
%
Although the FlowAct architecture is not contingent on a particular robotic platform, the experiments conducted in this study employ Pepper, a robot developed by Softbank Robotics. Consequently, the implementation details provided will be based on this platform while maintaining general applicability. 
Pepper is equipped with an array of four microphones, two loudspeakers, and three tactile sensors. Non-verbal communication is activated by LED clusters located in the eyes, on the shoulders, and around the ear speakers. Robot perception is aided by two 640x480 resolution cameras, strategically placed on the forehead and mouth, and an ASUS Xtion 3D sensor in one eye, essential for localization and navigation. Given its size, the forehead camera is ideal for HRI, aligned with the average human body height. Pepper runs under the NAOqi OS operating system, a GNU/Linux distribution based on Gentoo. For our experiment, NAOqi 2.5.5 is used, coupled with ROS\footnote{\url{https://wiki.ros.org/}}, so that the perceptions received and sent by NAOqi pass through topics (the message-passing framework implemented within ROS). These initial sensory data collected by the robot's primary sensors are defined as 'raw perceptions.'

\subsection{Perception refiner modules}
%
The FlowAct methodology encompasses the augmentation of the principal raw perceptions acquired by the agent, thereby facilitating the generation of a comprehensive scene analysis. This is accomplished through the implementation of perception refinement modules. These modules systematically process and enhance the initial dataset, resulting in a more sophisticated and insightful collection of perceptions, referred to as "refined perceptions." Subsequently, the refined perceptions are amalgamated with the raw perceptions and transmitted to the environment state tracker, which constructs a scene representation based on this synthesized data. 

In our implementation, the following modules are used:\\
\begin{itemize}
    \item \textbf{Voice activity detection}: detects speech segments in raw data, based on an adaptive threshold on energy of the sound signal.
    \item \textbf{Person tracker}: combines the location of the person and the extraction of body keypoint features from an implementation of Yolov7~\cite{yolov7_2022_wang} with a Deepsort algorithm~\cite{deepsort_2017} to track the identity of the people present in the images captured by the camera.
    \item \textbf{Gaze tracker}: uses the RT Gene's ROS package~\cite{rtgene_fischer_2018_ECCV} which transforms the image stream into various facial descriptors, gaze orientation, head pose, and the position of key facial points.
    \item \textbf{Speaker diarization}: is based on a temporal audio-visual fusion model for multi-user speaker diarization~\cite{speakerdia_2023}. The method identifies dominant speakers and tracks them over time by measuring the spatial coincidence between sound locations and visual presence. 
    \item \textbf{Interaction acceptance belief (IAB)}: infers the level of IAB~\cite{iab_2023} which commonly answers the question 'What are the chances of my interaction to be accepted by the targeted user' and is mainly based on the gaze of the user.
    \item \textbf{Automatic speech recognition (ASR):}  based on API calls sending the speech segments (detected by the VAD module) to recognizers (such as GoogleCloud speech recognition\footnote{\url{https://cloud.google.com/speech-to-text}} or a local OpenAI Whisper\footnote{\url{https://github.com/openai/whisper}} whenever user's privacy is at stake).
\end{itemize}

\subsection{Representation of the environment}
%
The EST controller is tasked with the construction of an accurate temporal representation of the scene. Consequently, it collects refined perceptual data, maintains a historical record, orchestrates the flow of perceptions, and delivers a comprehensive representation of the environment. Within this asynchronous gathering of perceptions, it systematically aligns and organizes these perceptions to synchronously deliver its representation of the environment.

EST is therefore responsible for associative memory~\cite{hri_mtt_prescott_2019}, a concept that defines the ability to link two initially unrelated elements, such as a name and an unknown face. In this context, we employ specific "Environment analysis modules," like person re-identification, which tracks users in the scene, or perception assigner, tasked with linking each detected user to a known or new identity.

In particular, in the task targeted in our experiments (Section 4), the environment representation focuses on a single subject type: users detected within images. Each user is associated with attributes such as their IAB value, the number of times they have been engaged, their corresponding utterances, and the agent's state. This latter includes information about its spatial position, current and past actions, most recent utterance, the user it is currently interacting with, and its current state (observation, engaged, return to its watchtower position, engaging user).

\subsection{Decision layer}
%
The agent's behavioral dynamics are orchestrated through the synergistic operation of the Action Planner controller and a suite of Action modules. The Action Planner serves as the cerebral core, permanently rendering decisions to either stay put, initiate interaction, return to a predefined (observation) position, or continue the current interaction. Decisions are seamlessly transmitted to the Action modules, each designed to spring into operation responsive to the delineated behavior. Building upon the framework established by Kanda \textit{et al.}~\cite{proactive_kanda_2009}, we conceptualize the outputs of the action planner as 'global behavior'. To adhere to this global behavior, the 'local behaviors' are relayed via action modules. These local behaviors are characterized by their more granular and atomic nature. For example, the action 'continue interaction' implies the speaking module to answer when it detects a user utterance. The engagement behavior in the tested system is governed by a series of rules based on the level of the IAB model in the refined perceptions, coupled with implicit engagement requests from the user (a raised hand for instance). The action planner frequency is $2$Hz.

For our experimental setup, we have implemented two pivotal Action modules:\\
\textbf{Moving Module}: capable of executing two distinct actions. Firstly, it engages with a specified individual by aligning the robot's orientation towards the target and proceeding to within 0.7 meters. Secondly, it navigates the robot back to its initial observational position. The system is equipped with internal states that ensure the robot remains stationed at its standby location and is engaged in a navigational sequence;\\
\textbf{Speaking Module}: This module regulates interaction with users, initiating a conversational cycle once engagement is achieved and the communication parameters are defined by the action modules. Responses are generated through an API call to a large language model (LLM), such as Vicuna\footnote{\url{https://lmsys.org/blog/2023-03-30-vicuna/}} configured as a conversational assistant, such as with the role-play zero-shot setting~\cite{roleplay2024}. Termination of the conversation is dependent upon the identification of specific keywords or user disengagement, including a 10-second interval of non-response. Additionally, this module updates internal state indicators, namely 'is speaking' and 'in a conversation', to reflect active speech and participation by the robot, thereby ensuring continuity in action planning decisions.

These modules collectively embody an agent's behavior as a sophisticated interaction of decision-making processes, supported by a feedback mechanism that enhances the agent's environmental awareness and adaptability progressively.

\section{\uppercase{Experiments}}
%
The experiments conducted to evaluate the implementation of FlowAct are executed in two distinct phases. The initial phase involves a controlled experiment within a laboratory setting, where individuals simulate the role of patients. This phase facilitates the testing and refinement of the system's functionalities, enabling the observation of interactions and responses of the proactive social robot within a controlled environment. The subsequent phase is carried out in a real-world setting, specifically within a hospital, involving actual patients. This phase seeks to assess the system under real-world conditions, considering unpredictable variables and the diversity of human interactions that may arise. This bifurcated approach will not only technically validate the system but also collect feedback on the user experience, ensuring the robot's capability to effectively perform its social role in a hospital context.
The modules are slightly modified between the laboratory experiment and the hospital experiment to ensure the anonymity of hospital patients; Google Speech-to-Text is replaced with the local version of Whisper.

\subsection{Laboratory experiments}
%
\subsubsection{Scenario}
%
To evaluate the effectiveness of the proposed model, we implemented the proactive interaction loop within a real-world setting, specifically designed to replicate a scenario where patients await their appointments in a hospital waiting room. To emulate the role of patients for this investigation, a cohort of diverse participants was enrolled, including 20 individuals, 14 men and 6 women, all in the age range [22, 52]. Only 3 of them were familiar with robotics. 

The participants, which consisted mainly of academics and students from our institution, gave their written consent to participate in this study. Furthermore, written informed consent was obtained from individuals for the publication of any identifiable images or data that could be included in publications related to the experiment.

The participants were positioned in the vicinity of the robotic entity (3-4 meters radius), either seated or standing. Thereafter, they were instructed to adopt one of the following behaviors, which are intended to represent different levels of signals for proactive engagement:\\
\begin{itemize}
    \item Engage in dialogues with individuals seated adjacent to them;
    \item Engage in active utilization of mobile devices, for instance, playing games or browsing the web;
    \item Exhibit a passive behavior, maintaining a stance of idleness and portraying a waiting state devoid of any particular engagement or activity;
    \item Display cues of interest and attentiveness towards the robotic agent;
    \item Initiate interaction with the robotic agent by seeking its attention or assistance, and requesting information, guidance, or support.
\end{itemize}

\begin{figure}[t!]
  \centering
  \includegraphics[scale=0.35]{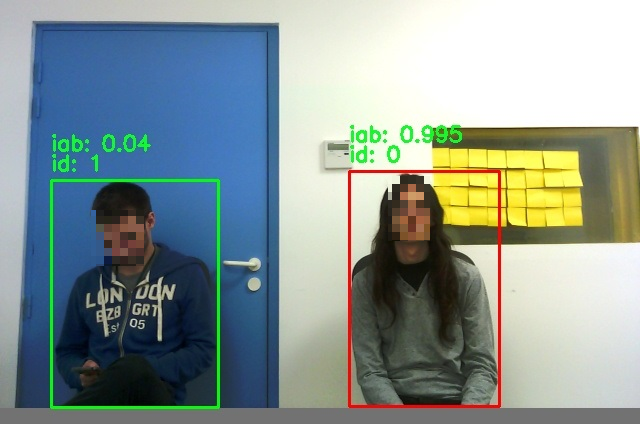}
  \caption{Screenshot of a scene, with anonymised faces, showing the IAB value for each person detected. Red box indicates a person available for engagement.}
  \label{figure:explab}
\end{figure}

Upon concluding an interaction, the robot returns to its observation position to initiate another engagement. The participants are then instructed to resume their designated scenarios, which may have been interrupted during the interaction, once the robot reverts to its original position. Participants were instructed to speak only during their interaction with the agent or when communicating with an individual seated next to them as part of the scenario. Their speaking turn during an interaction is indicated by an image displayed on the agent's tablet.

Each scenario encompasses a blend of passive behavior towards the agent, along with active behaviors such as showing interest or requesting interaction. A scenario is considered complete either when a predetermined time limit is reached or after each participant has been engaged by the robot.

\subsubsection{Evaluation}
%
To validate the functionality and usability of using FlowAct for proactive robotic interaction, we performed a thorough evaluation focusing on user experience to check the efficiency of the setup\footnote{All data collected during these evaluations can be made available for research purpose upon simple request to the main author.}. The experience is seen as a task in which the robot has to display a proactive engagement behavior toward the humans gathered in the room. The participants are informed about the task of the robot and rate the questionnaire accordingly to their posterior feelings.

The User Experience Questionnaire, as outlined by Finstad's~\cite{umux_finstad_2010} study, was employed to gauge users' interactions with the system. Each question represents a usability component evaluation of the system. Following the order of the questionnaire, we can measure effectiveness, satisfaction, overall quality, and efficiency. The Usability Metric for User Experience (UMUX), a concise and effective four-item Likert scale, was utilized for the subjective evaluation of the application's usability. This scale is strategically crafted to yield insights comparable to those derived from the more extensive 10-item system usability scale (SUS), ensuring a thorough and reliable assessment of user experience.

\subsubsection{Laboratory results}
%
The average UMUX score of approximately 71 suggests that, on a scale of 0 to 100, the overall usability and user experience of the system being evaluated are good. Regarding task performance, a significant majority of participants perceived the agent as effectively fulfilling its intended role, as in the example shown in Figure~\ref{figure:explab}. The robot exhibited its ability to participate in interactions within a hospital setting, demonstrating both reactive engagement in response to explicit user cues and proactive engagement prompted by the actor's exhibited interest in the agent.

Yet the general satisfaction with the system was mixed. Although most of the participants did not experience excessive frustration, notable instances of frustration were primarily attributed to the response latency of the agent, which typically ranged from 5 to 10 seconds. This delay was a consequence of the computational demand for the LLM-based conversational agent. In particular, participants with previous robotics experience expressed higher levels of frustration related to this latency.

In terms of usability, the consensus was that the system was user-friendly and did not require specific prerequisites for operation. During the experiments, it was observed that the score of efficiency decreased in situations where participants had to maintain eye contact with the agent for longer than anticipated (often exceeding 10 seconds) or when they were required to repeat themselves due to the robot's inability to comprehend their initial speech. In the course of the conducted trials, each experiment was successfully executed, demonstrating engagement and interaction with each participant. In particular, in two instances, the robot initiated interaction with individuals before they exhibited the reactive sign, typically a raised hand signaling the agent. The users perceived this preemptive interaction by the robot as proactive, as it occurred in response to their demonstrated interest prior to the conventional signal for engagement.

\subsection{Hospital experiments}
%
\subsubsection{Scenario and evaluation}
%
Subsequent to the laboratory experiments confirming operational performance, the experimental procedures were implemented in a hospital setting over a consecutive three-day period. A total of 11 patients were involved, 4 men and 7 women, resulting in a total of 13 recorded interactions. The mean age of the participants was 75 years, with ages ranging from 52 to 89 years. Of these 11 individuals, seven had prior experience interacting with robotic systems. All procedures conducted within the hospital setting were previously approved by the hospital's ethics committee.

The scenario at the hospital slightly differs from the laboratory setup. To evaluate system's usability and proper functioning of the observation-engagement-interaction loop, participants were asked, in turns, to first demonstrate interest in engaging the robot. If the robot did not engage proactively, participants were instructed to call it reactively (using a hand signal, for instance). Each scene involved one or two individuals positioned in front of the robot.

The evaluation setup for the hospital experiments differs slightly. The experiments are carried out using the SUS questionnaire~\cite{sus_brooke_1995}, preferred by our psychologist partners at the hospital. The SUS questionnaire is used to evaluate the perceptions of robot performance by participants in the context of interaction. The SUS questionnaire, similar to the previously used UMUX questionnaire, is a standardized tool that is used to assess the usability of a system. It consists of ten statements rated on a five-point Likert scale, ranging from "Strongly Disagree" to "Strongly Agree." This scale collects quantitative data on various aspects of the interaction, such as ease of use, perceived complexity, user confidence, and the learning curve. The scores are then converted into an overall score out of 100, making it easier to compare and interpret the results.

\subsubsection{Hospital results}
%
Despite our efforts and due to factors beyond our control related to the hospital context, this sample is still too small to draw firm conclusions from the collected observations. However, within the framework of this paper, it allows us to establish the operational context of the study and its practical implementation. Even though experimental sessions are needed to complement this very preliminary set, in the meantime, analyzing the current results may still offer some insights into the system's current state and the possibility of some immediate improvements before further collecting real-world interactions.

\begin{figure*}[ht!]
  \centering
  \includegraphics[scale=0.27]{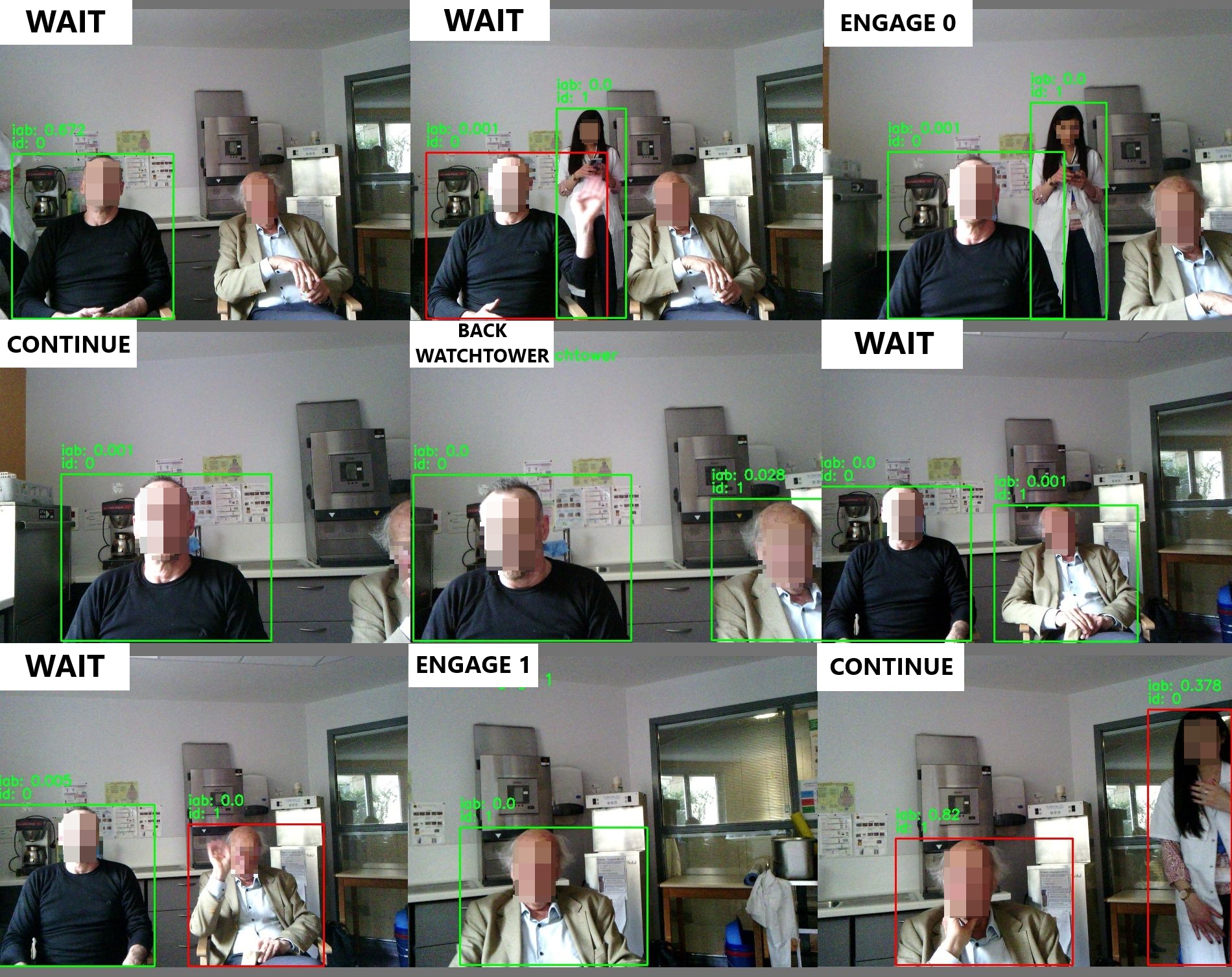}
  \caption{Example sequence of anonymised images showing patients' engagements the the hospital.}
  \label{fig:example_sequence_censored}
\end{figure*}

The average SUS score obtained from the evaluation is 59, indicating a moderate level of usability. According to the SUS scoring framework, scores close to 51 are considered "fair" or "so-so", while 71 are generally interpreted as "good"~\cite{Bangor2009DeterminingWI}. A score of 59, therefore, suggests that the system under evaluation presents notable usability challenges. This score implies that users may experience difficulties in interacting with the system, potentially affecting overall user satisfaction and engagement.

The average duration of the interactions is $4'2"$. The responses to the first question about general acceptance show that users are not yet ready to use the robot regularly. This is despite the fact that the perceived complexity is relatively low (Q2) and the ease of use is acknowledged (Q3). Patients feel capable of using the robot independently, but they believe that the integration of services could be improved. This sentiment is understandable given the observed slowness of the robot's movements and the response latency of the language module. Although inherent errors of the LLMs were present, the participants did not overly penalize inconsistencies in the dialogues (Q6). In general, participants find this human-robot interaction system simple and effective in terms of usability, capable of building trust with the patient, but they are not inclined to use it or face it frequently.

\subsection{Data analysis}
%
During the experiments, the detection of a high IAB value was prioritized to ensure patient engagement. However, if patients had difficulty engaging, they had the option to raise their hand to signal their willingness to engage. As a result, only 23\% of the interactions were initiated by detecting a high IAB value, while the remaining 77\% were initiated at the implicit request of the patients, through a hand gesture.

One primary reason was a too long activation time for the IAB (in laboratory settings IAB could take up to 5s before activation) and so participants were prone to display a reactive hand gesture signal immediately after showing (unanswered) signs of interest, such as gazing toward the agent. Also, hospital experiments took place in 3 different locations, with highly varying lighting conditions. So it made the IAB prediction more complex because the gaze orientation module, which feeds into the main module, is rather sensitive to light. Thus, after analyzing the data, we observed that successful engagement using proper IAB detection worked only in less than half of the interactions. Additionally, the person detection module was also challenged, as face detection is also sensitive to lighting conditions. These observations highlight the lack of robustness in the current IAB model.

Figure~(\ref{fig:example_sequence_censored}) illustrates a sequence of successive patient engagements. Following the image sequence from left to right and top to bottom, we observe that the patient on the right is not initially detected correctly, preventing any interaction at first. The robot starts in observation mode, where it detects a signal from the user to initiate engagement, in this case, a hand gesture. Once the conversation begins, the robot decides to continue until the dialogue module considers the conversation to be completed. The robot then returns to observation mode, waiting for a new signal to engage a patient in a new conversation. This pattern of 'observation post, engagement detection, interaction, return to observation post' represents the standard loop for the proactive objective of the agent in a hospital waiting room. Interestingly, the person in the white coat in the background is correctly detected but is not considered engageable in photos 2 and 3. This is likely due to her behavior on her phone, which does not mark her as a potential target. In the last photo, we observe that she moves closer to the engaged person and is then detected as engageable by her IAB value, which is consistent as she is directly looking at the agent while assisting the user.

\begin{table}[t!]
  \centering
  \caption{Example of a conversation patient-robot.}
  \label{tab:conversation_pepper}
 \makebox[\linewidth]{
    \begin{tabular}{cp{\dimexpr\linewidth-1.7cm}}
        \hline
        \textbf{Speaker} & \textbf{Dialogue turns} \\ \hline
        \textbf{Patient} & \textit{Hello, I'm X, and you?} \\ 
        \textbf{Pepper} & Hello, I am Pepper, the conversational robot of the hospital. How can I help you today? \\ 
        \textbf{Patient} & \textit{I am too early for my appointment, what should I do while I wait? }\\ 
        \textbf{Pepper} & You are early for your appointment. You can take a moment to relax and wait for the medical staff to take you to the consultation rooms at the scheduled time. This may take between 5 minutes and half an hour, depending on the number of consultations for the day. \\ 
        \textbf{Patient} & \textit{But how can I relax? What can I do to relax while I wait? }\\ 
        \textbf{Pepper} & You can take a moment to relax while waiting for your consultation by flipping through a book or listening to music. The medical staff will take you to the consultation rooms at the scheduled time. \\ 
        \textbf{Patient} & \textit{Can you play some music?} \\ 
        \textbf{Pepper} & I can suggest playing music, but as a robot, I don't have access to a music system. Would you like me to talk to you about music while you wait for your consultation? \\ 
        \textbf{Patient} & \textit{Sure.} \\ 
        \textbf{Pepper} & How are you today? \\ \hline
    \end{tabular}
 }
\end{table}

The dialogue in Table~\ref{tab:conversation_pepper} is an example of a conversation between a patient and the agent. Pepper demonstrates a good understanding of the user's requests and offers suggestions on how to pass the time while waiting for a medical appointment. When the user asks Pepper to play music, the robot explains its technical limitations (lack of access to a music system), but offers an alternative by suggesting they could talk about music. This response shows that the language model can handle technological limitations while maintaining an engaging interaction. However, it fails to follow up on the patient's confirmation of their desire to talk about music. This difficulty in maintaining a coherent dialogue beyond a few exchanges indicates that improvements are needed in how the chatbot handles conversation history. Despite knowing all previous exchanges, it still struggles to pursue a conversation to its logical conclusion.

During this trial, patient-robot conversations were difficult due to the low volume and insufficient clarity of the patients' speech. Sometimes a simple repetition was enough, but often the transcriptions differed significantly from the user's actual utterance. This complicated interactions with patients, highlighting the need to improve the sound capture system for effective use with elderly individuals.

\section{\uppercase{Conclusion}}
%
In summary, this research presents a system implementing the FlowAct architecture, a preliminary approach in the field of continuous perception-action systems within a robotic context for pro-active multimodal HRI. Central to this study is the development of a ROS architecture for a socially assistive robot, engineered to provide efficient services while dynamically and personally engaging users. The comprehensive evaluation conducted in a controlled environment not only validated the functionality of each module but also the efficacy of the architecture as a whole. Initial experimentation in simulated and real hospital settings has highlighted the system's proficiency in both proactive and responsive interactions with human participants.  Some conditions, however, will require an improvement of some individual perception and action modules to ensure greater public acceptance. 

\bibliographystyle{apalike}
{\small \footnotesize
\bibliography{example}}

%

\end{document}